\documentclass[hidelinks]{article}
\newcommand{\documentdate}{19 VI 2023}

\usepackage{a4wide,latexsym,graphicx,amsmath}
\usepackage{varioref}
\usepackage{xcolor}
\usepackage{ulem}
\usepackage{multirow}
\usepackage{hyperref}
\title{Divergence of the ADAM algorithm with fixed-stepsize: \\a
  (very) simple example}

\author{Philippe L. Toint\footnotemark[1]}

\newcommand{\beqn}[1]{\begin{equation}\label{#1}}
\newcommand{\eeqn}{\end{equation}}
\newcommand{\req}[1]{(\ref{#1})}

\newcommand{\tim}[1]{\;\; \mbox{#1} \;\;}

\newcommand{\ii}[1]{\{ 1, \ldots, #1 \}}

\renewcommand{\Re}{\hbox{I\hskip -2pt R}}
\newcommand{\smallRe}{\hbox{\footnotesize I\hskip -2pt R}}

\newcommand{\comment}[1]{}

\date{\documentdate}

\begin{document}

\maketitle

\renewcommand{\thefootnote}{\fnsymbol{footnote}}
\footnotetext[1]{Namur Center for Complex Systems (naXys),
  University of Namur, Namur, Belgium.
  Email: philippe.toint@unamur.be.}

\begin{abstract}
A very simple unidimensional function with Lipschitz continuous gradient is
constructed such that the ADAM algorithm with constant stepsize, started from the
origin, diverges when applied to minimize this function in the
absence of noise on the gradient. Divergence occurs irrespective of
the choice of the method parameters.
\end{abstract}

{\small
\textbf{Keywords:} ADAM algorithm, machine learning, deterministic
nonconvex optimization.
}

\section{Introduction}

This short note provides a new explicit example of failure
of the ADAM algorithm \cite{KingBa15}, one of the most popular training
methods in machine learning. Given the problem
\beqn{problem}
\min_{x \in \smallRe^n} f(x)
\eeqn
where $f$ is continuously differentiable function from $\Re^n$ into
$\Re$ with Lipschitz continuous gradient, and a starting iterate
$x_0$, the ADAM sequence of iterates is defined (see
\cite{ReddKaleKuma18}), for $i\in\ii{n}$ and $k\geq0$, by
the recurrences
\begin{align}
[m_k]_i    &= \beta_1 [m_{k-1}]_i + (1-\beta_1) [g_k]_i,\label{adam-m}\\
[v_k]_i    &= \beta_2 [v_{k-1}]_i + (1-\beta_2) [g_k]_i^2,\label{adam-v}\\
[x_{k+1}]_i &= [x_k]_i - \alpha\frac{ [m_k]_i}{\sqrt{[v_k]_i}},\label{adam-x}
\end{align}
where $[v]_i$
is the $i$-th component of the vector $v\in\Re^n$, $m_k$ is the $k$-th
``momentum'', $x_k$ is the $k$-th iterate, $g_k = \nabla_x^1f(x_k)$,
$\beta_1\in [0,1)$ is the momentum parameter and $\beta_2\in [0,1)$ 
is the ``forgetting'' parameter, and $\alpha>0$ is a (fixed)
steplength/learning-rate parameter. The recurrences \req{adam-m} and
\req{adam-v} are initialized by setting, for $i\in\ii{n}$, $[m_{-1}]_i = [g_0]_i$ and
$[v_{-1}]_i =  [g_0]_i^2$, respectively.
ADAM is intended to converge to find first-order points for problem \req{problem}, in
the sense that, for each $i\in\ii{n}$, $|[g_k]_i|$ should converge to zero when $k$
tends to infinity. In practice, this algorithm is most often used in a
stochastic context where the gradient $g_k$ is contaminated by noise
(typically resulting from sampling) and has generated a considerable
interest in the machine learning community.

Despite its widespread use, difficulties with this algorithm are not
new. In the noiseless (deterministic/full batch) case, obstacles for
proving convergence were in particular mentioned in
\cite{DefoBottBachUsun22}, essentially pointing out the possibility
that second-order terms in the Taylor's expansion of the objective
function could not vanish quickly enough. In
\cite[Theorem~1]{ReddKaleKuma18} an example of non convergence on a
convex function was produced in the online-learning stochastic
context, but this example crucially depends on the nonzero variance of
the noise. In a recent discussion at the June 2023 Thematic Einstein
Semester on Optimization and Machine Learning in Berlin, it was
suggested that, although likely, no explicit example of failure of
Adam with fixed stepsize was available for the deterministic case
(where the variance is zero). This prompted the author to produce the
(very simple) one which is, for the record, detailed in the next
section.  We note that an again convex but deterministic example had
already been provided in the comprehensive analysis of ADAM's
behaviour (with decreasing stepsize) detailed in \cite{Zhanetal22b}
(see Propositions~3.3 and E1). This analysis describes conditions
which delineate a region strictly included in $[0,1)^2$ such that ADAM
with parameters $\beta_1$ and $\beta_2$ chosen in this region
generates a diverging sequence on this example. In contrast, the
simple example we are about to discuss is nonconvex and applies to the entire
$[0,1)^2$, but requires constant stepsize.  It can therefore be seen
as complementing the analysis of \cite{Zhanetal22b}.

\section{The example}

\begin{figure}[ht] 
\centerline{\includegraphics[width=7cm]{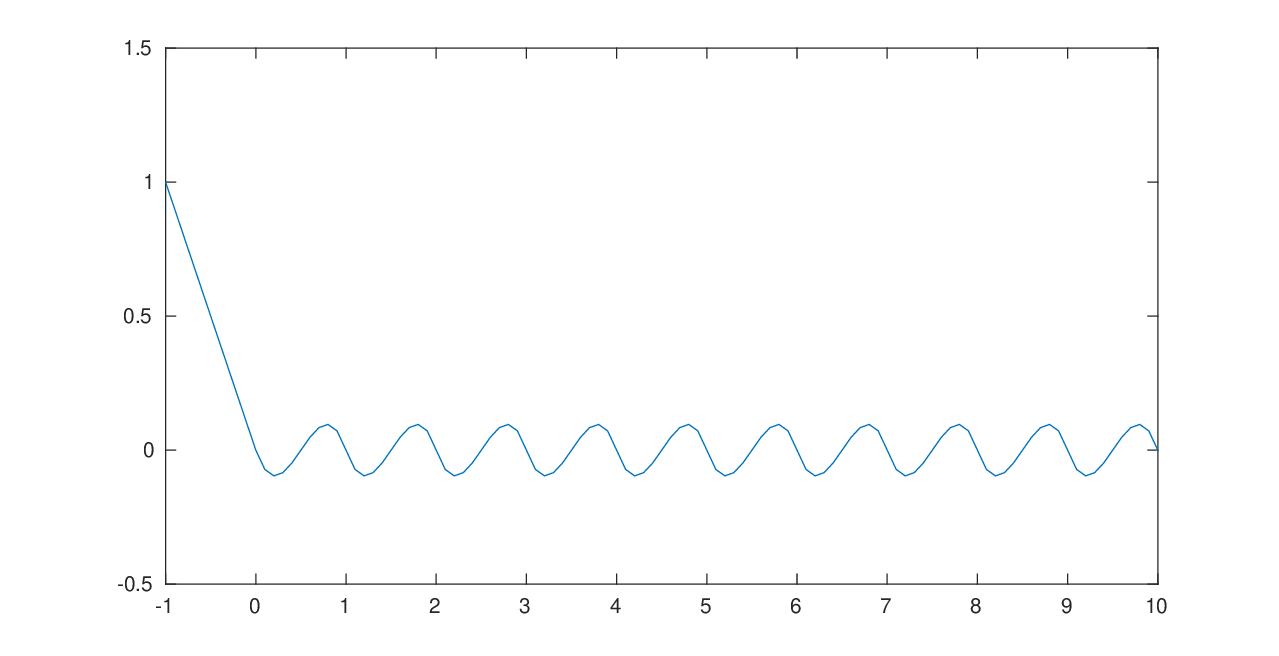}}
\caption{\label{slowadam} The shape of $f(t)$ for small values of $|t|$}
\end{figure}

To show that the ADAM algorithm may fail to converge on nonconvex
functions with Lipschitz gradient, we will exhibit an example in
dimension one, which we construct in two stages.  We first define
sequences of iterates, together with associated function and gradient values
which remain constant throughout the iterations. We next verify that
these sequences may be considered as generated by applying the ADAM
algorithm to a nonconvex objective function with Lipschitz gradient.
(Since the example is unidimensional, we omit the component indices
($i$) if what follows.)
For $k\geq 0$, let the sequence of function values and gradients be
defined by
\beqn{grads}
f_k  = 0 \tim{ and } g_k = - 1,
\eeqn
and the sequence of (potential) iterates be defined (from \req{adam-m}-\req{adam-x})
by
\begin{align}
m_k    &= \beta_1 m_{k-1} + (1-\beta_1) g_k = -1,\label{1Dadam-m}\\
v_k    &= \beta_2 v_{k-1} + (1-\beta_2) g_k^2 = 1,\label{1Dadam-v}\\
x_{k+1} &= x_k - \alpha\frac{ m_k}{\sqrt{v_k}} = x_k + \alpha,\label{1Dadam-x}
\end{align}
where we used \req{grads} to derive the last equality in \req{1Dadam-m} and \req{1Dadam-v}.
Thus
$\sum_{j=0}^k\beta^{k-j} \leq 1/(1-\beta)$  for $\beta\in(0,1)$ imply that
\beqn{slower}
 s_k = x_{k+1}-x_k=\alpha,
\eeqn
for $k\geq0$ and  $x_k$ tends tp infinity.
We now show that there exists a (nonconvex) univariate function $f_1$
defined on $\Re^+$ with Lipschitz continuous gradient such that $f_k=
f(x_k)=0$ and $g_k=\nabla_x^1f_1(x_k)=-1$ for all $k \geq 0$. Indeed,
a simple Hermite interpolation calculation based of these conditions
yields that, for all $t\geq0$, 
\beqn{f1def}
f_1(t) = -(t-x_{k(t)})
+\frac{3}{s_{k(t)}}\,(t-x_{k(t)})^2 -
\frac{2}{s_{k(t)}^2}\,(t-x_{k(t)})^3,
\eeqn
where $k(t)$ is such that $t\in [x_k,x_{k+1}]$.
We may then define
\[
f(t)   = \left\{\begin{array}{ll}
f_1(t) &\tim{if } t\geq0,\\
-t     &\tim{if } t< 0,
\end{array}\right.
\]
so that $f(t)$ is well-defined on the whole of $\Re$, has Lipschitz
continuous gradient and is such that the ADAM algorithm
\req{1Dadam-m}-\req{1Dadam-x} applied on $f$ starting from $x_0=0$ generates iterates with
$|g_k|= 1$ for all $k\geq0$. We thus conclude that the ADAM algorithm
fails to converge on this particular instance of problem
\req{problem}. A graph of $f(t)$ for $t\in [-1,10]$, $\beta_1=\beta_2 
= 0.9$ is shown in Figure~\ref{slowadam}. 
One also verifies that the Lipschitz constant 
on the interval $[x_k,x_{k+1}]$ is given by
\[
L_k = \sup_{t\in (x_l,x_{k+1})}|\nabla_t^2f(t)| = \frac{6}{s_k}
\]
so that, using \req{slower},
\[
L = \max_{k\geq0}L_k
=\frac{6}{\alpha}
\]
Moreover, defining $T_k(s) = f_k + g_ks$, it results from \req{grads},
\cite[Theorem~A.9.2]{CartGoulToin22}, \req{slower} and the
inequalities
\[
|f_{k+1}-T_k(s_k)| = s_k \leq \frac{1}{\alpha} s_k^2
\tim{and}
|g_{k+1}- \nabla_s^1T_k(s)| = |-1 +1 | \leq \frac{1}{\alpha} s_k
\]
that $f(t)$ is bounded below by a constant only depending on $\alpha$.
As a consequence, we see that, \textit{for
any fixed $(\beta_1,\beta_2) \in[0,1)^2$ and $\alpha>0$, there exist
unidimensional functions with Lipschitz continuous gradient whose
gradient's Lipschitz constant is as small as $6/\alpha$, which is
bounded below by a constant only depending on $\alpha$ and for which
the ADAM algorithm \req{1Dadam-m}-\req{1Dadam-x} starting from $x_0=0$
generates iterates with constant nonzero gradients (therefore failing
to converge)}.

Since our example is unidimensional and since ADAM is defined
componentwise, the same conclusion obviously applies irrespective of
$n$, the problem dimension. Indeed divergence in a single component implies
divergence on the whole space.

Our result thus extends that of \cite{Zhanetal22b} in that it includes 
methods for arbitrary $(\beta_1,\beta_2)\in(0,1)^2$ but fixed
stepsize. Note that $|\nabla_t^1f(t)|$ is bounded 
by $L$ for all $t\in \Re$, again at variance with the
example of this reference.
 
Observe that our conclusions would also hold if we had fixed $g_k$ to 
another negative constant (we can multiply $f$ by this constant) or if, instead of \req{adam-x},
we had considered
\[
[x_{k+1}]_i = [x_k]_i - \frac{\alpha\,[m_k]_i}{\sqrt{\epsilon+[v_k]_i^2}},
\tim{ or }
    [x_{k+1}]_i = [x_k]_i - \frac{\alpha\,[m_k]_i}{\epsilon+\sqrt{[v_k]_i^2}},
\]
where $\epsilon$ is a small positive constant, but they do not apply
in the more realistic situation where stepsizes $\alpha_k\rightarrow
0$ are used (as is for instance the case in
\cite[Proposition~1.1]{Zhanetal22b}, where $\alpha_k$ is a multiple of $1/\sqrt{k}$). 
We finally note that we have chosen a constant zero value for $f_k$ in
order to simplify our bounds, but that it is also possible to choose
$f_{k+1} > f_k$ (leading to an monotonically increasing sequence of
function values) without qualitatively affecting our
conclusion, although this leads to a larger value of $L$.

\section*{\footnotesize Acknowledgement}

{\footnotesize
Thanks to  Satyen Kale, Omri Weinstein, Alena
Kopani\v{c}\'akov\'a and Serge Gratton for interesting exchanges.


}
\end{document}